\documentclass{svproc}

\usepackage{amsmath}
\usepackage{amssymb}
\usepackage{cite}
\usepackage{comment}
\usepackage{graphicx}
\usepackage{siunitx}
\usepackage{xcolor}
\allowdisplaybreaks

\usepackage{soul}
\usepackage{array}
\usepackage{float}
\usepackage{mathtools}
\usepackage{multirow}
\usepackage{wasysym}
\usepackage{svg}
\usepackage{import}

\makeatletter
\let\NAT@parse\undefined
\makeatother
\usepackage[allcolors=blue,colorlinks=true]{hyperref}

\usepackage{caption}
\usepackage{subcaption}

\renewcommand{\baselinestretch}{0.96}

\makeatletter%change heading spacings
\renewcommand{\section}{\@startsection{section}{1}{\z@}{1ex plus 1ex minus 0.5ex}%
	{0.9ex plus 1.2ex minus 0ex}{\normalfont\normalsize\centering\scshape}}%
\makeatother

\usepackage{microtype}

\title{Smooth real-time motion planning based on a cascade  dual-quaternion screw-geometry  MPC}
\author{Ainoor Teimoorzadeh, Frederico Fernandes Afonso Silva, Luis F.C. Figueredo and Sami Haddadin} 
\authorrunning{Ainoor Teimoorzadeh et al.}
\titlerunning{  }
 \institute{
Chair of Robotics and Systems Intelligence,\\ 
Munich Institute of Robotics and Machine Intelligence, \\
Technical University of Munich, DE-80992 Munich, Germany,\\
\email{ainoor.teimoorzadeh@tum.de}
}
\begin{document}
\maketitle

\global\long\def\dq#1{\underline{\boldsymbol{#1}}}%

\global\long\def\quat#1{\boldsymbol{#1}}%

\global\long\def\mymatrix#1{\boldsymbol{#1}}%

\global\long\def\myvec#1{\boldsymbol{#1}}%

\global\long\def\mapvec#1{\boldsymbol{#1}}%

\global\long\def\dualvector#1{\underline{\boldsymbol{#1}}}%

\global\long\def\dual{\varepsilon}%

\global\long\def\dotproduct#1{\langle#1\rangle}%

\global\long\def\norm#1{\left\Vert #1\right\Vert }%

\global\long\def\mydual#1{\underline{#1}}%

\global\long\def\hamilton#1#2{\overset{#1}{\operatorname{\mymatrix H}}\left(#2\right)}%

\global\long\def\hamiquat#1#2{\overset{#1}{\operatorname{\mymatrix H}}_{4}\left(#2\right)}%

\global\long\def\hami#1{\overset{#1}{\operatorname{\mymatrix H}}}%

\global\long\def\tplus{\dq{{\cal T}}}%

\global\long\def\getp#1{\operatorname{\mathcal{P}}\left(#1\right)}%

\global\long\def\getd#1{\operatorname{\mathcal{D}}\left(#1\right)}%

\global\long\def\swap#1{\text{swap}\{#1\}}%

\global\long\def\imi{\hat{\imath}}%

\global\long\def\imj{\hat{\jmath}}%

\global\long\def\imk{\hat{k}}%

\global\long\def\real#1{\operatorname{\mathrm{Re}}\left(#1\right)}%

\global\long\def\imag#1{\operatorname{\mathrm{Im}}\left(#1\right)}%

\global\long\def\imvec{\boldsymbol{\imath}}%

\global\long\def\vector{\operatorname{vec}}%

\global\long\def\mathpzc#1{\fontmathpzc{#1}}%

\global\long\def\cost#1#2{\underset{\text{#2}}{\operatorname{\text{C}}}\left(\ensuremath{#1}\right)}%

\global\long\def\diag#1{\operatorname{diag}\left(#1\right)}%

\global\long\def\frame#1{\mathcal{F}_{#1}}%

\global\long\def\ad#1#2{\text{Ad}\left(#1\right)#2}%

\global\long\def\spin{\text{Spin}(3)}%

\global\long\def\spinr{\text{Spin}(3){\ltimes}\mathbb{R}^{3}}%

\global\long\def\unitquatgroup{\ensuremath{\text{Spin}(3)}}%

\global\long\def\unitquatset{\mathcal{S}^{3}}%

\global\long\def\dualquatset{\mathbb{H}\otimes\mathbb{D}}%

\global\long\def\unitdualquatgroup{\text{Spin}(3)\ltimes\mathbb{R}^{3}}%

\global\long\def\unitdualquatset{\dq{\mathcal{S}}}%

\global\long\def\closedballset{\ensuremath{\mathbb{B}}}%

\global\long\def\complexset{\ensuremath{\mathbb{C}}}%

\global\long\def\quatset{\ensuremath{\mathbb{H}}}% 
\global\long\def\purequatset{\ensuremath{\mathbb{H}_p}}% 

\global\long\def\realset{\ensuremath{\mathbb{R}}}%

\global\long\def\rationalset{\ensuremath{\mathbb{Q}}}%

\global\long\def\integerset{\ensuremath{\mathbb{Z}}}%

\global\long\def\naturalset{\ensuremath{\mathbb{N}}}%

\global\long\def\SE#1{\ensuremath{SE(#1)}}%

\global\long\def\SO#1{\ensuremath{SO(#1)}}%

\global\long\def\SU#1{\ensuremath{SU(#1)}}%

\newcommand{\fred}[1]{\textcolor{blue}{#1}} % Fred's modifications
\newcommand{\fredcomment}[1]{\textcolor{red}{#1}} % Fred's comments

%%%%%%%%%%%%%%%%%%%%%%%%%%%%%%%%%%%%%%%%%%%%%%%%%%%%%%%%%%%%%%%%%%%%%%%%%%%%%%%%%%%%%%%%%%%%%%%%%%%%%%%%%%%%%    A B S T R A C T %%%%%%%%%%%%%%%%%%%%%%%%%%%%%%%%%%%%%%%%%%%%%
\begin{abstract}
This paper investigates the tracking problem of a smooth coordinate-invariant trajectory using dual quaternion algebra. The proposed architecture consists of a cascade structure in which the outer-loop MPC performs real-time smoothing of the manipulator's end-effector twist while an inner-loop kinematic controller ensures tracking of the instantaneous desired end-effector pose. Experiments on a $7$-DoF Franka Emika Panda robotic manipulator validate the proposed method demonstrating its application to constraint the robot twists, accelerations and jerks within prescribed bounds.
\end{abstract}
%%%%%%%%%%%%%%%%%%%%%%%%%%%%%%%%%%%%%%%%%%%%%%%%%%%%%%%%%%%%%%%%%%%%%%%%%%%%%%%%%%%%%%%%%%%%%%%%%%%%%%%%%% I N T R O D U C T I O N %%%%%%%%%%%%%%%%%%%%%%%%%%%%%%%%%%%%%%%%%%%%%%%
\section{INTRODUCTION}
Robotic manipulators are gaining broad acceptance in a wide range of applications, varying from manufacturing to assistive care. As the robotic applications of serial manipulators grow, so does the complexity of the environment and the conditions in which they operate. This brings extra challenges into the planning and control formulation, as constraints on the end-effector should be satisfied during motion. For instance, carrying a bottle of water may require holding the cup straight while allowing free motion in the Cartesian space---regardless of the reference frame used to describe the motion, e.g., reference frame from a camera or robot base. Furthermore, when sharing the environment with humans, the robot should prioritize coordinate invariant, easy to demonstrate and real-time smooth movements, while ensuring the imposition of safety constraints \cite{Khatib1986, Marinho2018, Marinho2019, kirschner2022expectable}.

To address this challenge, this work proposes a reactive planning approach based on the sate-of-the-art coordinate-invariant real-time path planning. The approach is built on the screw geometry of motion along side a model-predic\-tive control and kinodynamic constraints in the task space.
 Separately, screw motion and model-predictive control are well-developed areas within the control and robotics communities. The former has been largely studied with a multitude of applications \cite{Dimentberg1965, Waldron1972, Agrawal1987, Murray1994, Pennock1992, Wohlhart1995, Aspragathos1998, Selig2004, Cibicik2020, Muller2021a, Silva2022,figueredo2021robust}, and more recently integrated into real-time path planning approaches given its coordinate invariance properties \cite{Ge1994, Vochten2015, laha2022coordinate}, whereas the latter has experienced a surge of results in the past decades, particularly within field-robotics \cite{liu2017path}. Nevertheless, only a few studies have explored solutions falling in the intersection of both fields, and to the best of the authors’ knowledge, none have been shown to satisfy the inherent translation and rotation coupling that describes rigid body motions in SE$(3)$, as well as screw geometry with constraint and coordinate-invariance properties necessary for real-world constraint satisfaction.

Reactive motion and path planners for real-world applications often rely on dynamics or control systems with convergence and stabilization properties, such as \cite{schaal2003control,haddadin2011dynamic}, or on geometric constraint satisfaction through interpolation methods \cite{laha21cooperative}. Although the former provides stabilization features and allows for kinodynamic planning, they often lack formal guarantees to satisfy geometric constraints and have poor explainability of the resulting trajectory, which is only locally optimized. In contrast, interpolation approaches are easy to deploy, interpret, and generalize from demonstrations; that is, a simple demonstration can be used to guide a path planner, as seen in \cite{laha21cooperative}. Therefore, they are often preferred in industry and service robotics. However, care must be taken to ensure a proper group structure when performing interpolation and addressing the geometric constraints. Several results still decouple translation and rotation components---and some even consider Euler angles, which are widely known to be singular and non-representative for motions---which lead to poor results and dependency on the reference frame.\footnote{Coordinate invariance should always be addressed. Otherwise, one may even have different results in trajectory tracking simply by changing features within an object topology, e.g., visual tracking of different edges in a mesh or a point cloud.}  Readers are referred to \cite{Allmendinger2018,sarker2020screw,laha21p2p,laha2022coordinate}.  In this study, our planner produces paths through screw linear interpolation (ScLERP) \cite{sarker2020screw} that implicitly maps—and therefore satisfies—all geometric constraints embedded in the single demonstration. 

Notwithstanding, defining smooth continuous, $C_1$ or $C_2$ trajectories through demonstration or interpolation whilst satisfying the inherent geometric constraints is not trivial. As one includes additional velocity and acceleration constraints, following the trajectory and keypoints becomes challenging. Allmendinger et al. \cite{Allmendinger2018} proposes a parameter selection for a dual-quaternion $C_1$ screw linear interpolation but the approach is fairly limited to special cases. 
In contrast, recent quadratic-programming and model-predictive control solutions are well-suited to ensure constraint satisfaction with fewer parameters. 
Particularly, 
model predictive control (MPC) formulations allow the control of constrained multiple-input multiple-output nonlinear systems with respect to an optimal criteria. Similarly to $H_2$ and $H_\infty$ strategies \cite{fei2006tracking} and to quadratic-programming-based controllers \cite{todorov2006optimal}, MPC approaches explore the solution of an optimal control problem. However, differently from the former, MPCs have a finite time horizon that enables the online solution of the optimal control problem \cite{mayne2000constrained, pereira2021nonlinear}, making it suitable for dynamic trajectory tracking within cluttered environments \cite{kamel2017linear}. The ongoing advancements in the underlying theoretical framework has evolved MPC into a reliable control technique capable of offering stable, robust, constraint-compliant controllers, and computationally feasible solutions for both linear and nonlinear systems \cite{ferramosca2008mpc}. Moreover, the predictive capability of MPCs lead to enhanced trajectory tracking performance by effectively handling disturbances and generating smooth control signals \cite{mayne2000constrained}. 
However, most approaches have a decoupled treatment of the translational and rotational components of the mechanism \cite{rao1999steady}. Thus, increasing the number of equations, the overall complexity of the system, and often leading to the reference-frame dependency. For instance, Pereira et al. \cite{pereira2021nonlinear} used a $SE(3)$ representation of a quadrotor UAV and of the obstacles present in a cluttered environment. Based on that, the authors proposed a nonlinear model predictive control strategy for the execution of aggressive maneuvers. However, albeit free from the representation singularities inherent from the extraction of Euler angles from rotation matrices, both the dynamic model of the UAV and the control formulation present decoupled equations for the attitude and the position of the system, given in $SO(3)$ and in $\mathbb{R}^{3}$, respectively.

This paper presents the integration of a screw-interpolation strategy that satisfies path and geometric constraints within a coordinate-invariant manner with a dual-quaternion algebra MPC that constraints twists, accelerations and jerks within prescribed bounds. Dual quaternions provide a unified representation of the angular and linear components with strong geometrical meaning while being free of representational singularities, more compact, and having lower computational costs than homogeneous transformation matrices \cite{Adorno2011}. The proposed architecture is a cascade structure in which the outer-most MPC controller smooths the coordinate-invariant trajectory of the end-effector twist in real-time while the inner-loop kinematic controller ensures tracking of the instantaneous desired end-effector pose. 

We validate the proposed method in experiments on a $7$-DoF Franka Emika Panda robotic manipulator including constraints on the robot twists, accelerations and jerks within prescribed bounds.

%%%%%%%%%%%%%%%%%%%%%%%%%%%%%%%%%%%%%%%%%%%%%%%%%%%%%%%%%%%%%%%%%%%%%%%%%%%%%%%%%%%%%%%%%%%%%%%%%%%%%%%%%%%%%%%%%%%%%%%% PROBLEM DEF %%%%%%%%%%%%%%%%%%%%%%%%%%%%%%%%%%%%%%%%%%%%%%%%%%%%%%%%%%%%%%%%%%%%%%%%%%%%%%%%%%%%%
\section{Problem Formulation}\label{Sec:ProbDef}

In this section, we provide an overview of the definitions and fundamentals of the planning problem. Firstly, we review the core concepts related to the algebra of dual quaternions (DQ) and the properties of a screw-linear theory based first-order interpolation. These will build the backbone of the proposed planning scheme. 

%%%%%%%%%%%%%%%%%%%%%%%%%%%%%%%%%%%%%%%%%%%%%%%%%%%%%%%%%%%%%%%% MATH %%%%%%%%%%%%%%%%%%%%%%%%%%%%%%%%%%%%%%%%%%%%%%%%%%%%%%%%%%%%%%%%
\subsection{Mathematical  Preliminaries\label{sec:mathematical-preliminaries}}

Dual quaternions \cite{Selig2005} are elements of the set
\begin{align*}
\mathcal{H} & \triangleq\{\quat h_{\mathcal{P}}+\dual\quat h_{\mathcal{D}}:\quat h_{\mathcal{P}},\quat h_{\mathcal{D}}\in\mathbb{H},\,\dual\neq0,\,\dual^{2}=0\},
\end{align*}
where $\mathbb{H}\triangleq\{h_{1}+\imi h_{2}+\imj h_{3}+\imk h_{4}\::\:h_{1},h_{2},h_{3},h_{4}\in\mathbb{R}\}$
is the set of quaternions, in which $\imi$, $\imj$ and $\imk$ are
imaginary units with the properties $\imi^{2}=\imj^{2}=\imk^{2}=\imi\imj\imk=-1$
\cite{Selig2005}. Addition and multiplication of dual quaternions
are analogous to their counterparts of real and complex numbers. One
must only respect the properties of the dual unit $\dual$ and imaginary
units $\imi,\imj,\imk$.

The subset $\dq{\mathcal{S}}=\left\{ \dq h\in\mathcal{H}:\norm{\dq h}=1\right\} $ is the subset
of unit dual quaternions, where $\norm{\dq h}=\sqrt{\dq h\dq h^{*}}=\sqrt{\dq h^{*}\dq h}$,
with $\dq h^{*}$ being the conjugate of $\dq h$ \cite{adorno2017robot}. Under the  multiplication operation, this subset defines the group $\unitdualquatgroup$ which double covers $\SE{3}$ \cite{Selig2005}. Any arbitrary rigid body transformation can be represented by the unit dual quaternion $\dq{x} \in \unitdualquatgroup$,  
\begin{equation}
    \dq x=\quat r+\dual\left(1/2\right)\quat p\quat r, 
    \label{eq:dq}
\end{equation}
where $\quat p=\imi x+\imj y+\imk z \in \purequatset$ represents the Cartesian position $\left(x,y,z\right)$ within the set of pure quaternions, i.e., $\mathbb{H}_{p}\triangleq\left\{ \quat h\in\mathbb{H}:\real{\quat h}=0\right\}$, where the real component is null, i.e., $\real{h_{1}+\imi h_{2}+\imj h_{3}+\imk h_{4}}\triangleq h_{1}$. 
The rotation  $\quat r=\cos\left(\phi/2\right)+\quat n\sin\left(\phi/2\right)$ 
is defined within the unit-quaternion group, $\unitquatgroup$, in which $\phi\in[0,2\pi)$ is the rotation
angle around the rotation axis $\quat n$. Notice the rotation axis $\quat n$ is a unitary pure quaternion, that is, $\quat n\in\mathbb{H}_{p}\cap\mathbb{S}^{3}$
with 
% $\mathbb{H}_{p}\triangleq\left\{ \quat h\in\mathbb{H}:\real{\quat h}=0\right\} $,
% where $\real{h_{1}+\imi h_{2}+\imj h_{3}+\imk h_{4}}\triangleq h_{1}$,
% and 
$\mathbb{S}^{3}=\left\{ \quat h\in\mathbb{H}:\norm{\quat h}=1\right\} $. For further details of the unit-dual quaternion representation see 
\cite{Aspragathos1998,Selig2005,2011_Adorno_THESIS,adorno2017robot,2016_Figueredo_PhDThesis, Silva2022thesis}.

The set $\mathcal{H}_{p}=\left\{ \dq h\in\mathcal{H}:\real{\dq h}=0\right\} $
of pure dual quaternions is used to represent twists and wrenches,
which are represented in different coordinate systems using the adjoint
operator $\mathrm{Ad}:\dq{\mathcal{S}}\times\mathcal{H}_{p}\to\mathcal{H}_{p}$.
For instance, consider the twist $\dq{\xi}^{a}\in\mathcal{H}_{p}$
expressed in frame $\frame a$ and the unit dual quaternion $\dq x_{a}^{b}$
that represents the rigid motion from $\frame b$ to $\frame a$.
The same twist is expressed in frame $\frame b$ as
\begin{equation}
\dq{\xi}^{b}=\ad{\dq x_{a}^{b}}{\dq{\xi}^{a}}=\dq x_{a}^{b}\dq{\xi}^{a}\left(\dq x_{a}^{b}\right)^{*}.\label{eq:adj_transf}
\end{equation}

Furthermore, it is critical to highlight that from a differential geometry perspective, the Lie group associated to $\unitdualquatgroup$ is defined within a  differentiable Riemannian manifold \cite{2017_Busam_Birdal_Navab_ICCVW}. As a direct consequence, Riemannian metrics based on a collection of inner products on the tangent space at $ \unitdualquatgroup$ can be assigned to the manifold \cite{Book:Boothby:2002,2017_Busam_Birdal_Navab_ICCVW,laha2022coordinate}. These Riemannian metrics define the length of paths along the manifold \cite{1995_Park_JMD_ASME}, and therefore allow us to define  minimum curve lengths, i.e., geodesics, see  \cite{1995_Park_JMD_ASME,Book:Boothby:2002,Zacur2014b,2009_Sachkov_JMS} for further information. In such manifolds, actions in the geodesics can be expressed by means of the exponential map $\exp _{\dq x} : \mathcal{T}_{\dq x} \unitdualquatgroup  \rightarrow \unitdualquatgroup  $. The $\exp _{\dq x} $ locally maps a vector in the tangent space $\mathcal{T}_{\dq x}  \unitdualquatgroup $ (at  $\dq x  \in  \unitdualquatgroup$)\footnote{The tangent space at $\dq x$ is built by the collection of vectors whose inner products with $\dq x$ is null---that is the orthogonal vector space to $\dq x$.} to a point on the manifold following the geodesic through $\dq x $ \cite{2017_Busam_Birdal_Navab__ArXiv}. The inverse mapping (from manifold to tangent space at the point $\dq x$) is the logarithm map $ \log _{\dq x} : \unitdualquatgroup \rightarrow  \mathcal{T}_{\dq x} \unitdualquatgroup  $.

The mappings $\exp _{\dq x} $ and $\log _{\dq x}$ are non-trivial to obtain. A solution is to compute them by parallel transport \cite{2017_Busam_Birdal_Navab_ICCVW,2013_Lorenzi_Pennec__IJCV}. The parallel transport exploits the exponential function that maps vectors from the tangent space (at the identity) to the manifold \cite{Zacur2014b},
\begin{align}
    \exp _{\dq x}(\dq y)   & = \dq x \exp ( \dq x^\ast \dq  y  ),    \nonumber \\
    \log _{\dq x}(\dq z)   &  = \dq x \log ( \dq x^\ast \dq z  ) ,   
    \label{eq:parallelTransport}
\end{align}
where  $\dq  z \in \unitdualquatgroup  $ and $\dq  y $  is defined in the tangent space at $\dq x$---notice that $\dq y$ is not a unit DQ. The $\exp$ and $\log$  maps from the tangent space, at the identity, i.e., $\mathcal{T}_{\dq 1}~ \unitdualquatgroup$ are given by the dual vector representing the axis of screw motion and the dual angle containing both the translation length and the angle of rotation, see further details in \cite{2017_Busam_Birdal_Navab_ICCVW,2017_Busam_Birdal_Navab__ArXiv,sarker2020screw,2012_Wang_Han_Yu_Zheng_JMAA,laha2022coordinate,adorno2017robot}.  

%%%%%%%%%%%%%%%%%%%%%%%%%%%%%%%%%%%%%%%%%%%%%%%%%%%%%%%%%%%%%%%% OVERVIEW OF THE PROBLEM %%%%%%%%%%%%%%%%%%%%%%%%%%%%%%%%%%%%%%%%%%%%%%%%%%%%%%%%%%%%%%%%
\subsection{Overview of the Problem\label{sec:sec:problem overview}}

In this work, we are interested in the design of a real-time motion planning solution that considers geometric constrains from prescribed keypoints in a coordinate-invariant fashion, while ensuring smooth movements and addressing twist, acceleration and jerk constraints in real-time. The proposed motion generation scheme takes as prior knowledge any $n$-number of keypoints in task-space, 
\begin{equation}
    \mathcal{K} =\{ \dq{k}_1, \dq{k}_2, \dots, \dq{k}_\ell, \dots \dq{k}_n  \}, \ \  \dq{k}_\ell \in \unitdualquatgroup.
    \label{eq:keypoints}
\end{equation}

These keypoints implicitly embed the desired task-space constraints and range. In this way, following a screw-linear interpolation, our planner also ensures that the resulting sequence of rigid body transformations from interpolation satisfies the observed task-space constraints in pose, i.e., orientation and translation as well. Notwithstanding, in case either the prescribed or resulting twists, acceleration or jerks are not feasible, the real-robot system would fail in deployment. To ensure additional constraint satisfaction, not only instantaneous, but rather along the trajectory, we propose a cascade approach with a model-predictive control system that takes the dual-quaternion algebra and the mapping to the tangent space of the prescribed poses in the path into account. This leads to a smooth motion planner satisfying the below problem definition. 

\ 

\noindent \textbf{Problem Definition: }
Given a set of $n$-number of keypoints in the task-space, $\mathcal{K}$, with $n\geq2$, find a trajectory from $\dq k_0 $ to $\dq k_n$ such that 
\begin{enumerate}
    \item The implicit constraints within $\mathcal{K}$ are satisfied as close as possible;
    \item Motion generation is achieved in real-time with an additional constraint satisfaction and smooth motion regarding twists, acceleration and jerk constraints.
\end{enumerate}

%%%%%%%%%%%%%%%%%%%%%%%%%%%%%%%%%%%%%%%%%%%%%%%%%%%%%%%%%%%%%%%%%%%%%%%%%%%%%%%%%%%%%%%%%%%%%%%%%%%%%%%%%%%%%%%%%%%%%%%% MPC MODELING %%%%%%%%%%%%%%%%%%%%%%%%%%%%%%%%%%%%%%%%%%%%%%%%%%%%%%%%%%%%%%%%%%%%%%%%%%%%%%%%%%%%%
\section{ScLERP-MPC: A Motion Planner based on Screw-linear Interpolation \& Model-Predictive Control}\label{MPC-modeling}

This section presents an integrated screw-linear interpolation with a model-predictive control solution to address the problems designed in the problem definition. From the desired set of $n$ keypoints in the task-space, $\mathcal{K}$, we first need an initial path planning structure going through along the desired setpoints. 

The ScLERP \cite{sarker2020screw} explores the screw-linear interpolation that connects any two points through the geodesic prescribed in the previous section. Given $\dq x_{a}$ and $\dq x_{b}$, the resulting path should be given by $\dq x(\tau):[0,1]\rightarrow\unitdualquatgroup$ with $\dq x(0)=\dq x_{a}$ and $\dq x(1)=\dq x_{b}$. The process starts by mapping $\dq x_{b}$ following the geodesic on $\unitdualquatgroup$ through $\dq x_{a}$ onto the tangent space at $\dq x_{a}$. In other words, it obtains a $\mathcal{T}_{\dq x _a} \unitdualquatgroup$ corresponding to the geodesic direction of $\dq x_{b}$ w.r.t. $\dq x_{a}$. Hence,
\begin{equation}
    \log_{\dq x_{a}}(\dq x_{b})=\dq x_{a}\log(\dq x_{a}^{\ast}\dq x_{b}),    
    \label{eq:parallel_for_x1_to_x2}
\end{equation}
where the mapping $\exp_{\dq x_{1}}$ and $\log_{\dq x_{a}}$ is computed using the parallel transport \eqref{eq:parallelTransport}, and defines the tangent space of a Riemannian manifold---a vector space.
From the geodesic path in the tangent space, one can linearly interpolate points from $\log_{\dq x_{a}}(\dq x_{a})$ towards $\log_{\dq x_{a}}(\dq x_{b})$, as 
% \begin{equation}
$
    (\log_{\dq x_{a}}(\dq x_{b})-\log_{\dq x_{a}}(\dq x_{a}))\tau +\log_{\dq x_{a}}(\dq x_{a}),  
$
with  $\log_{\dq x_{a}}(\dq x_{a}){=}0$.   
Hence,
using the parallel transport \eqref{eq:parallelTransport} to map the vector  in $\mathcal T _{\dq x_{a}}{\unitdualquatgroup}$
back to the $\unitdualquatset$ manifold (following the geodesics along $\dq x_{a}$) gives way to  
\begin{align}
\dq x(\tau)= & \exp_{\dq x_{a}}\left(\dq x_{a}\log(\dq x_{a}^{\ast}\dq x_{b})   \tau \right)\nonumber \\
% = & \dq x_{1}\exp\left(\dq x_{1}^{\ast}\dq x_{1}\log(\dq x_{1}^{\ast}\dq x_{2}) \tau  \right)\nonumber \\
= & \dq x_{a}\exp\left(\log(\dq x_{a}^{\ast}\dq x_{b})   \tau  \right).\label{eq:sclerp_equation}
\end{align}
%-----------------------------------

Following \eqref{eq:sclerp_equation}, the prescribed discrete path linearly-scaled along the geodesic between two keypoints $\dq{x}_a$ and $\dq{x}_b$ can be derived as \cite{sarker2020screw}
\begin{equation}\label{eq:x_d}
    \dq{x} = \text{ScLERP}(\dq x_a,\dq x_b;\tau) = \dq x_a (\dq x_a^{-1} \dq x_b)^\tau, 
\end{equation}
with $\tau \in [0~1]$ defined within equally spaced values.   
Notice the ScLERP function \eqref{eq:x_d} is the same as the one derived in \eqref{eq:sclerp_equation}. This can be shown by geometrical exponential \cite{adorno2017robot,article:1996_Kim_Kim_Shin__JVCA},  
% and similarly by means of  geometrical first order integration \cite{Adorno2017} 
and from the scaling of the dual rotation angle about the screw axis---hence the name \cite{DQBlending}. 
Furthermore, the ScLERP interpolation allows for the 
coordinate-invariant interpolation which is not possible when decoupling orientation and translation \cite{grassmann2018smooth, sarker2020screw}, as detailed in \cite{Allmendinger2018}.\footnote{Similar interpolation scheme nonetheless could also be derived from  $\SE{3}$, 
and other covering groups that satisfy left-invariance and are based on non-minimal representation of rigid displacements.  
Hence, it is by no means restricted to the choice of $\unitdualquatgroup$. %---it will lead to similar results from any coordinate-invariant transformation based on non-minimal representations of rigid body displacements. 
Still, a matrix-based solution is non-attractive due to the additional computational cost---that can possibly restrict real-time implementation---and due to the efficiency, compactness and intuitiveness of $\unitdualquatgroup$ which can depict wrenches, twists, geometric primitives, constraints and its tangent space with the same algebra.}   

The resulting screw interpolation can be used to connect all keypoints from \eqref{eq:keypoints}. The resulting connected path from the coordinate-invariant ScLERP interpolation \eqref{eq:x_d} through $\dq{k}_\ell$ to $\dq{k}_{\ell{+}1}$ within $\mathcal{K}$, $\ell={1,\dots,  n{-}1}$, results in a discrete set of desired poses  $\dq{x}_d $.  The desired twist between the discrete points can be either user-defined or follow a $C_0$ path. In this case, the prescribed reference twist, $\dq{\xi}_r \in \mathcal{H}_{p}$, is given in a way to describe the geodesic path within the given time-step $i$,  
\begin{equation}\label{eq:xi_r}
    \dq{\xi}_r [i]= \frac{2}{\tau}\log (x_d[i]x_d^\ast[i-1]).
\end{equation}

Notwithstanding the result trajectory is $C_0$, and hence might not be feasible for the robot system to execute. Thus, for our framework we integrate a  discrete MPC  to improve smoothness and moreover ensure the twist, acceleration and jerk constraints in the task-space are satisfied. The discrete MPC optimizes the future control trajectory within the finite control horizon $n_c \in \mathbb{N}$ in the prediction horizon. To track the desired trajectory, we consider the system as a double integrator, that is $\myvec u[i]=\vector_{6}\left(\ddot{\dq{\xi}_{r}}\right)\in\mathbb{R}^{6}$, in which the operator $\vector_{6}:\mathcal{H}\to\mathbb{R}^{6}$ maps the coefficients of a pure dual quaternion into a sixth-dimensional vector.\footnote{
{Given $\dq h=\imi h_{2}+\imj h_{3}+\imk h_{4}+\dual\left(\imi h_{6}+\imj h_{7}+\imk h_{8}\right)$, $\vector_{6}\dq h=\begin{bmatrix}h_{2} & \cdots & h_{8}\end{bmatrix}^{T}$.}} The state space equations are given by
\begin{equation}\label{eq:state space}
    \begin{aligned}
        \begin{bmatrix}
            \dot{\xi}_{\mathrm{r}} [i]\\
            \ddot{\xi}_{\mathrm{r}} [i]
        \end{bmatrix}  =  
        \underbrace{\begin{bmatrix}
            \mymatrix{0}_{6 \times 6} & \mymatrix{I}_{6 \times 6}\\
            \mymatrix{0}_{6 \times 6} & \mymatrix{0}_{6 \times 6}
        \end{bmatrix}}_{\mymatrix{A}_m}
        \begin{bmatrix}
            \xi_{\mathrm{r}} [i]\\
            \dot{\xi}_{\mathrm{r}} [i]
        \end{bmatrix}  + 
        \underbrace{  \begin{bmatrix}
            \mymatrix{0}_{6 \times 6}\\
            \mymatrix{I}_{6 \times 6}
        \end{bmatrix}}_{\myvec{B}_m} & \myvec u[i],\\
         \xi_{\mathrm{eff}} [i]=
         \underbrace{\begin{bmatrix} 
        \mymatrix{I}_{6 \times 6} & \mymatrix{0}_{6 \times 6}\end{bmatrix}}_{\mymatrix{C}_m}\begin{bmatrix}
        \xi_{\mathrm{eff}}[i]\\
        \dot{\xi}_r[i]
        \end{bmatrix},\\
    \end{aligned}
\end{equation}
where $\ensuremath{\dot{\xi}_{\mathrm{eff}}}=\vector_{6}\left(\ensuremath{\dot{\dq{\xi}}_{\mathrm{eff}}}\right)$ and $\ensuremath{\ddot{\xi}_{\mathrm{eff}}}=\vector_{6}\left(\ensuremath{\ddot{\dq{\xi}}_{\mathrm{eff}}}\right)$ are, respectively, the first and the second order time derivatives of the end-effector twist and $\mymatrix{I}_{6 \times 6}, \mymatrix{0}_{6 \times 6} \in \mathbb{R}^{6 \times 6}$ are zero and identity matrices.

Applying the backward difference operator, the augmented state vector $ \xi[i+1]=\begin{bmatrix}\Delta\Xi_{\mathrm{r}}^{T}[i+1] & \xi_{\mathrm{eff}}[i+1]\end{bmatrix}^{T}\in\mathbb{R}^{18} $, with $\Delta \Xi[i+1] = \begin{bmatrix} \Delta \xi_{\mathrm{r}} [i+1] & \Delta \dot{\dq{\xi}}_{\mathrm{r}} [i+1] \end{bmatrix}^T \in \mathbb{R}^{12}$, can be described by
\begin{equation} \label{eq:sate1}
    \xi[i+1]=
    \left[\begin{array}{cc}
    \mymatrix A_{m} & \mymatrix 0_{12\times6}\\
    \mymatrix C_{m}A_{m} & \mymatrix I_{6\times6}
    \end{array}\right]\left[\xi[i]\right]+\left[\begin{array}{c}
    \mymatrix B_{m}\\
    \mymatrix C_{m}\mymatrix B_{m}
    \end{array}\right]\left[\Delta\mymatrix u[i]\right]
\end{equation}
and also
\begin{equation}\label{eq:sate2}
    \xi_{\mathrm{eff}}[i]=\underbrace{\left[\begin{array}{cc}
\mymatrix 0_{6\times12} & \mymatrix I_{6\times6}\end{array}\right]}_{C}\left[\begin{array}{c}
\Delta\Xi_{\mathrm{r}}\\
\xi_{\mathrm{eff}}[i]
\end{array}\right]
\end{equation}

The control objective is to find the sequence of incremental control efforts $\Delta \myvec U $ over the control horizon, defined as
\begin{equation}\label{eq:delta_U}
    \Delta \myvec U = \begin{bmatrix}
 \Delta \myvec u^{T}[i] & &  \Delta \myvec u^{T}[i+1] & &\cdots & & \Delta \myvec u^{T}[i+n_c-1]
\end{bmatrix}^T,
\end{equation} 
such that $\Delta \myvec U \in \mathbb{R}^{6n_c}$ is the solution of minimizing the formulation of the cost function based $\mymatrix L$ on the Laguerre equations \cite{wang2009model} taken as 
\begin{equation}\label{eq:cost}
   \underset{\Delta \myvec U}{\min}\,\mymatrix L(\Delta\myvec U)=\left\Vert \myvec{\xi}_{s}-\myvec Y\right\Vert _{Q_{\mathrm{mpc}}}+\left\Vert \Delta\myvec U\right\Vert _{R_{\mathrm{mpc}}},
\end{equation}
subject to
\begin{equation}\label{eq:constraints}
\begin{cases}
\begin{array}{c}
\dot{\myvec \vartheta}_{min} \leq \Sigma \myvec U \leq \dot{\myvec \vartheta}_{max}              \\
\ddot{\myvec \vartheta}_{min} \leq \myvec U \leq \ddot{\myvec \vartheta}_{max}              \\
\dddot{\myvec \vartheta}_{min} \leq \Delta \myvec U \leq \dddot{\myvec \vartheta}_{max}
\end{array}\end{cases}
\end{equation}
in which $\begin{bmatrix}
 \dot{\myvec \vartheta}_{min} & \dot{\myvec \vartheta}_{max}
\end{bmatrix}\in \mathbb{R}^{12}$, $\begin{bmatrix}
 \ddot{\myvec \vartheta}_{min} & \ddot{\myvec \vartheta}_{max}
\end{bmatrix}\in \mathbb{R}^{12}$, and $\begin{bmatrix}
 \dddot{\myvec \vartheta}_{min} & \dddot{\myvec \vartheta}_{max}
\end{bmatrix}\in \mathbb{R}^{12}$ determine the limits in the Cartesian space for the admissible linear and angular velocities, accelerations, and jerks in the task pace respectively and $\myvec \xi^T_{s} \in \mathbb{R}^{6 \times n_p}$ is the vector that contains the information about the set points at the sampling time, 
\begin{equation}\label{eq:1_1_8}
    \myvec \xi^T_{s} = \underbrace{\begin{bmatrix}
     1 & & 1 & & \cdots & & 1
    \end{bmatrix}}_{n_{p}} \xi[i].
\end{equation}

The predicted output signal $\myvec Y $ in the equation \eqref{eq:cost}, which satisfies the boundary conditions on the upper and lower velocity, acceleration, and jerk bounds at the task space of the $n$-DoF manipulator, is the solution of \eqref{eq:sate1} and \eqref{eq:sate2}, described as 
\begin{equation}{\label{eq:1_1_7}}
    \myvec Y = \begin{bmatrix}\xi_{\mathrm{eff}}[i+1|i] & & \xi_{\mathrm{eff}}[i+2|i] & & \cdots & & \xi_{\mathrm{eff}}[i+n_{p}|i]\end{bmatrix} = \mymatrix F\xi[i]+\mymatrix{\phi}\Delta\myvec U,
\end{equation}
where $\xi_{\mathrm{eff}}[i+n_{p}|i]$ is the predicted twist at $i+n_{p}$ given the current
plant information at sampling time $i$.
\begin{equation}
    \begin{aligned}
        & \myvec F =\begin{bmatrix}
        C A & 
        CA^{2} &
        CA^{3} &
        \cdots &
        CA^{n_{p}}
        \end{bmatrix}^T \in \mathbb{R}^{6 \times (18*n_p)}, 
        \\& \myvec \phi =\begin{bmatrix}
        CB & O &  O &  \cdots & O\\
        CAB & CB &  O & \cdots & O\\
        CA^2B & CAB & CB & \cdots & O\\
        \vdots & \vdots & \vdots & \vdots & \vdots\\
        CA^{n_{p}-1}B & CA^{n_{p}-2}B & CA^{n_{p}-3}B &\ldots & CA^{n_{p}-n_{c}}B
        \end{bmatrix} \in \mathbb{R}^{(6*n_p) \times (6*n_c) }
    \end{aligned}
\end{equation}
\begin{comment}
    \begin{equation}
        \begin{aligned}
        & F=\begin{bmatrix}
        C A & 
        CA^{2} &
        CA^{3} &
        \cdots &
        CA^{n_{p}}
        \end{bmatrix}^T \in \mathbb{R}^{2m \times (q*n_p)}, 
        \\& \phi=\begin{bmatrix}
        CB & O &  O &  \cdots & O\\
        CAB & CB &  O & \cdots & O\\
        CA^2B & CAB & CB & \cdots & O\\
        \vdots & \vdots & \vdots & \vdots & \vdots\\
        CA^{n_{p}-1}B & CA^{n_{p}-2}B & CA^{n_{p}-3}B &\ldots & CA^{n_{p}-n_{c}}B
        \end{bmatrix} \in \mathbb{R}^{(q*n_p) \times (2m*n_c) },
        \end{aligned}
    \end{equation}
\end{comment}
will result in a predicted sequence of the state vectors 
\begin{equation}
    \begin{bmatrix}
        \xi[i+1|i] &    \xi[i+2|i] & \cdots & \xi[i+n_p|i]
    \end{bmatrix}.\label{eq:predicted_sequence} 
\end{equation}

From \eqref{eq:predicted_sequence}, the smoothed desired pose is obtained by
\begin{equation}\label{eq:error}
    \dq{x}_{d}[i] = \mathrm{exp}(\frac{k \dq{\xi}[i+1]}{2})\dq{x}_{d}[i-1],
\end{equation}
$k$ is the integration step.
    
Finally, the error between the current end-effector pose and the desired pose $\dq{x}_{d}[i]$ is defined as 
\begin{equation}
    \dq e [i+1] = 1 - \dq x_{d}^ \ast[i+1] \dq x_{\mathrm{eff}}[i].\label{eq:pose_error}
\end{equation} 

Defining $u_{\dot{q}} = \dot{\myvec q}$, and taking into account \eqref{eq:pose_error}, consider the following control law to ensure the closed-loop stability of the system \cite{laha2022coordinate}
\begin{equation}\label{eq:inner_control_law}
    u_{\dot{q}}=-(\hami -_{8}\left(\dq x_{d}[i]\right)\mymatrix C_{8}\myvec J)^{\dagger}\mymatrix K\vector_{8}\left(\dq e[i+1]\right)
\end{equation}
where $\mymatrix K$ is a positive definite gain matrix, $\hami -_{8}:\mathcal{H}\to\mathbb{R}^{8\times8}$ is the Hamilton operator, such that $\vector_{8}\left(\dq h_{1}\dq h_{2}\right)=\hami -_{8}\left(\dq h_{2}\right)\vector_{8}\dq h_{1}$, and the matrix $\mymatrix C_{8} \in \mathbb{R}^{8\times8}$ is defined as $\mymatrix C_{8}\triangleq\mathrm{diag}\left(\left[\begin{array}{cccccccc}
1 & -1 & -1 & -1 & 1 & -1 & -1 & -1\end{array}\right]\right)$ and $\myvec J \in \mathbb{R}^{8 \times 7}$ is the geometric jacobian  \cite{Adorno2011}. 

\begin{figure}
\begin{centering}
\scriptsize
\def\svgwidth{0.94\columnwidth}%
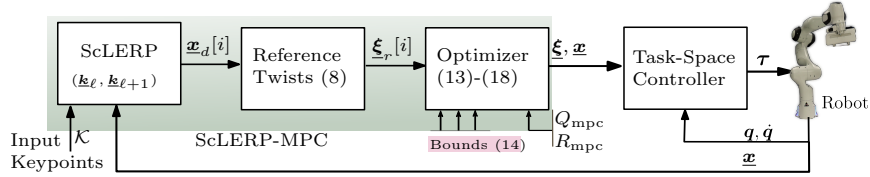%
\par\end{centering}
\caption{Schematic block diagram of the ScLERP-MPC.}
\label{fig:block-diagram}
\end{figure}

Fig.\,\ref{fig:block-diagram} presents the conceptual block scheme of the overall proposed control architecture.

%%%%%%%%%%%%%%%%%%%%%%%%%%%%%%%%%%%%%%%%%%%%%%%%%%%%%%%%%%%%%%%%%%%%%%%%%%%%%%%%%%%%%%%%%%%%%%%%% SIMULATIONS AND RESULTS %%%%%%%%%%%%%%%%%%%%%%%%%%%%%%%%%%%%%%%%%%%%%%%%%%%%%%%%%%%%%%%%%%%%%%%%%%%%%%%%%%%%%%%%%%%%%%%%%%%%%%%%%%%%%%%%%%%%%%%%%%%%%%%%%%%%%%
\section{Experimental Results} \label{Simulation and Results}
To validate the proposed ScLERP-MPC formulation, we performed two sets of tests. In the first one, we performed simulations in CoppeliaSim \cite{rohmer2013v} using a $7$-DoF Franka Emika Robot to demonstrate the capabilities of the MPC on imposing the desired constraints. The second validation was done through experiments on the real platform.

\subsection{Experimental setup}\label{experimental-setup}
To get the solution of the cost function in equation \eqref{eq:cost}, the following optimization problem is defined as the quadratic program given by
\begin{equation}
   \underset{\Delta \mymatrix U}{\min}\,\mymatrix L(\Delta\myvec U)=\frac{1}{2}\Delta\mymatrix U^{T}(\myvec{\phi}^{T}\mymatrix Q_{mpc}\myvec{\phi}+\mymatrix R_{mpc})\Delta\mymatrix U+(\myvec{\xi}_{s}-\mymatrix F\xi)\mymatrix Q_{mpc}\Delta\mymatrix U
\end{equation}
subject to,
\begin{equation}\label{eq:constraints_general}
    \mymatrix W \Delta \mymatrix U \leq \mymatrix V,
\end{equation}
where
\begin{equation*}
    \mymatrix W  =
    \begin{bmatrix}
        \mymatrix W_1 & \mymatrix W_2 & \mymatrix W_3
    \end{bmatrix}^T,
\end{equation*}
in which
\begin{equation*}
    \mymatrix W_1 = \mymatrix W_2 = \mymatrix W_3 = 
    \begin{bmatrix}
      -\mymatrix I & \mymatrix 0 & \mymatrix 0 & \cdots & \mymatrix 0 & \mymatrix 0 \\
       \mymatrix I & \mymatrix 0 & \mymatrix 0 & \cdots & \mymatrix 0 & \mymatrix 0 \\
       \mymatrix 0 & \mymatrix I & \mymatrix 0 & \cdots & \mymatrix 0 & \mymatrix 0 \\
       \mymatrix 0 & -\mymatrix I & \mymatrix 0 & \cdots & \mymatrix 0 & \mymatrix 0 \\
       \vdots & \vdots & \vdots & \cdots & \vdots  & \vdots\\
      \mymatrix 0 & \mymatrix 0 & \mymatrix 0 & \cdots & \mymatrix 0 & -\mymatrix I \\
       \mymatrix 0 & \mymatrix 0 & \mymatrix 0 & \cdots & \mymatrix 0 & \mymatrix I \\
    \end{bmatrix} \in \mathbb{R}^{12n_{c} \times 6n_{c}}
\end{equation*}
$I,O \in \mathbb{R}^{6 \times 6}$ are the identity and zero matrix respectively, and 
\begin{equation*}
    \mymatrix V = \begin{bmatrix}
 \mymatrix V_1 &  \mymatrix V_2 & \mymatrix V_3
\end{bmatrix}^T,
\end{equation*}
in which,
\begin{align*}
    \mymatrix V_{1} & = 
    \left[\begin{array}{ccccccc}
        -\dddot{\myvec \vartheta}_{min} & \dddot{\myvec \vartheta}_{max} & -\dddot{\myvec \vartheta}_{min} & \dddot{\myvec \vartheta}_{max} & \ldots & -\dddot{\myvec \vartheta}_{min} & \dddot{\myvec \vartheta}_{max}
    \end{array}\right]^{T}\in\mathbb{R}^{12n_{c}}, \\
    \mymatrix V_{2} & =
    \left[\begin{array}{ccccccc}
        -\ddot{\myvec \vartheta}_{min} & \ddot{\myvec \vartheta}_{max} & -\ddot{\myvec \vartheta}_{min} & \ddot{\myvec \vartheta}_{max} & \ldots & -\ddot{\myvec \vartheta}_{min} & \ddot{\myvec \vartheta}_{max}
    \end{array}\right]^{T} \\ & \quad \: - 
    \mymatrix W_{2}
    \left[\begin{array}{ccccccc}
        \ddot{\myvec \vartheta}[i-1] & \ddot{\myvec \vartheta}[i-1] & \ddot{\myvec \vartheta}[i-1] & \ddot{\myvec \vartheta}[i-1] & \ldots & \ddot{\myvec \vartheta}[i-1] & \ddot{\myvec \vartheta}[i-1]
    \end{array}\right]^{T} \in\mathbb{R}^{12n_{c}}, \\
    \mymatrix V_{3} &= 
    \left[\begin{array}{cc}
        \mymatrix{\phi}^{\dagger}(-\dot{\myvec \vartheta}_{min}+\mymatrix F\xi[i]) & \mymatrix{\phi}^{\dagger}(\dot{\myvec \vartheta}_{max}-\mymatrix F\xi[i])
    \end{array}\right. \\ & \quad \quad \quad \quad \quad \quad \quad \quad \quad \quad
    \left.\begin{array}{ccc}
        \ldots & \mymatrix{\phi}^{\dagger}(-\dot{\myvec \vartheta}_{min}+\mymatrix F\xi[i]) & \mymatrix{\phi}^{\dagger}(\dot{\myvec \vartheta}_{max}-\mymatrix F\xi[i])
    \end{array}\right]\in\mathbb{R}^{12n_{c}}.
\end{align*}
Furthermore, the upper and lower jerk, acceleration, and velocity constraints were selected to respect the limits of the Franka Emika Panda.\footnote{https://frankaemika.github.io/docs/control\_parameters.html}

% The desired initial $\dq x_1\in\dq{\mathcal{S}}$ and final $\dq x_2\in\dq{\mathcal{S}}$ poses were arbitrarily chosen and the simulations were performed in a control horizon of $n_c = 10$ over the prediction range $n_p = 50$. The sampling rate for the MPC controller was $9ms$ while the low level controller was running at $1kHz$. The stop criteria for the experiment was $\dq e  <= \mathrm{tol}$, with $\mathrm{tol} \in \mathbb{R}$ empirically defined.
The control horizon was chosen as $n_c = 10$, and the prediction range as $n_p = 50$. The stop criteria for the tests was $\dq e \leq \mathrm{tol}$, with $\mathrm{tol} \in \mathbb{R}$ empirically defined. For the experiments on the real platform, the sampling rate for the MPC controller was $9ms$ while the low level controller was running at $1kHz$.

\subsection{Simulations}\label{experimental-setup}
For the simulations, we selected an interpolated trajectory that would intentionally force the robot outside the desired constraints to demonstrate that the proposed ScLERP-MPC formulation can ensure their enforcement.

\begin{figure}
\begin{centering}
\scriptsize
\def\svgwidth{0.75\columnwidth}%{0.94\columnwidth}%
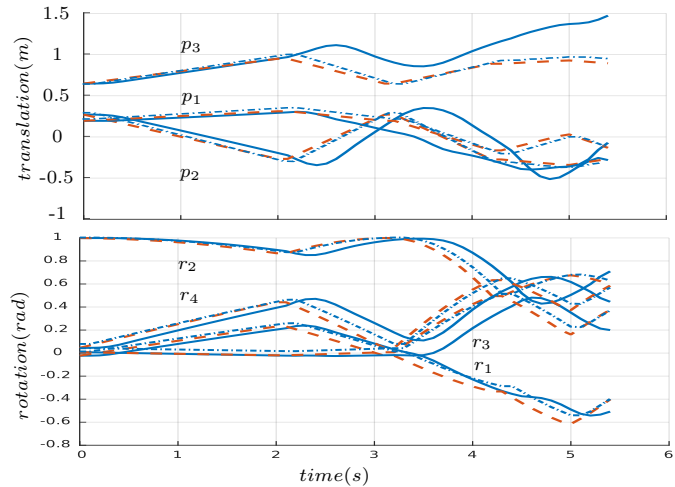%
\par\end{centering}
\caption{%Translational(\textit{top}) and angular (\textit{bottom}) components of the end-effector %twist 
% for the simulation. 
Trajectories for the simulated scenario. \textit{Solid blue} curves correspond to the SCLERP-MPC resulting trajectory whereas %values read from the robot,
\textit{Dashed blue} correspond a system %to the values read from the robot 
without kinodynamic constraints. %, whereas 
The \textit{Dashed red} curves refer to the reference.}
\label{fig:sim:rot:trans}
\end{figure}

\begin{figure}
\begin{centering}
\scriptsize
\def\svgwidth{0.84\columnwidth}%
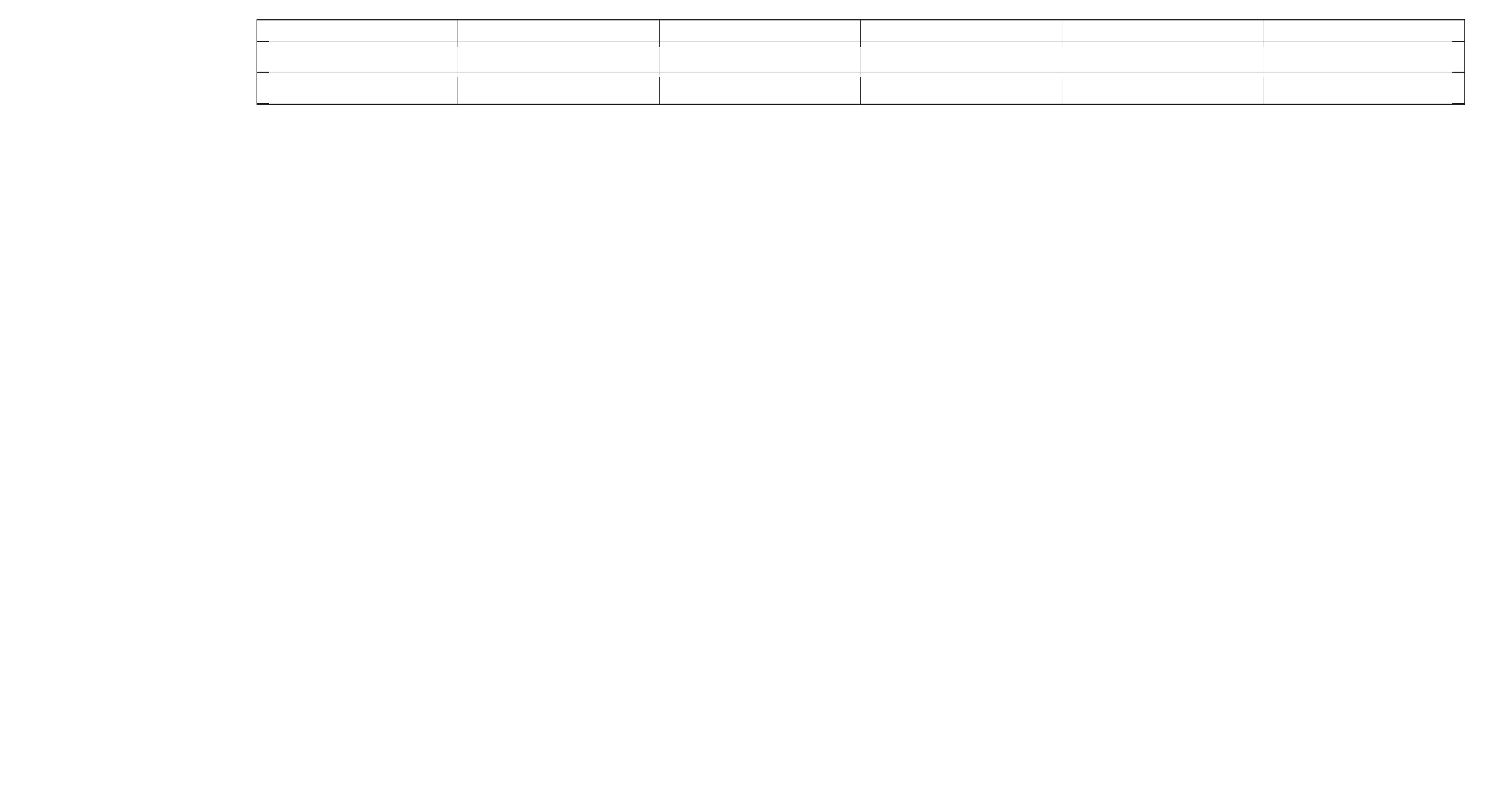%
\par\end{centering}
\caption{Twist trajectory for the simulation scenario. \textit{Solid blue} curves correspond to the SCLERP-MPC resulting trajectory whereas the \textit{Dashed red} curves refer to the reference. The shadowed areas depict picks of acceleration and jerk above the robot limits.}
\label{fig:sim:twists}
\end{figure}

 Fig.~\ref{fig:sim:twists} presents the end-effector twists, whereas Fig.~\ref{fig:sim:rot:trans} shows the angular and linear components of the end-effector pose. We can see that the executed trajectory presents a delay since the robot cannot violate the constraints of accelerations, and jerks. %as shown in Fig.\ref{fig:sim:acceleration}, and Fig.\ref{fig:sim:jerk}.
This behaviour is also observed in the resulting trajectory, Fig.~ \ref{fig:sim:rot:trans}.

\begin{figure}
\begin{centering}
\scriptsize
\def\svgwidth{0.84\columnwidth}%
 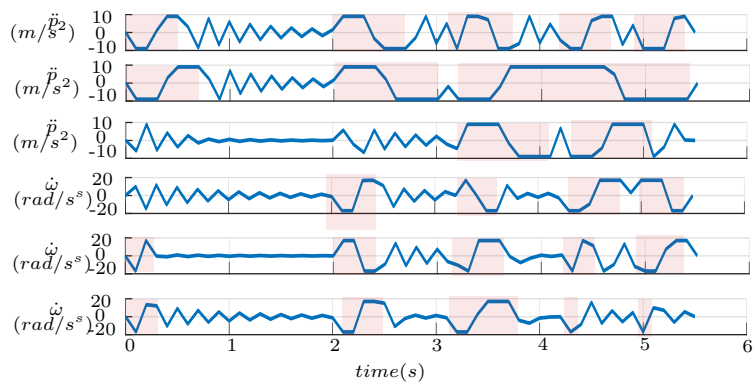%
 \par\end{centering}
 \caption{Acceleration trajectory for the simulation scenario. \textit{Solid blue} curves correspond to the SCLERP-MPC resulting trajectory. The shadowed areas depict picks of acceleration and jerk above the robot limits.}
 \label{fig:sim:acceleration}
\end{figure}

\subsection{Experiments}\label{experimental-setup}
For the experiments, the robot followed a trajectory obtained through the ScLERP method given two used-defined initial and final end-effector poses. The constraints imposed in the jerks, accelerations, and velocities followed the manufacturer recommendations to ensure safety of operation.

\begin{figure}
\begin{centering}
\scriptsize
\def\svgwidth{0.8\columnwidth}%
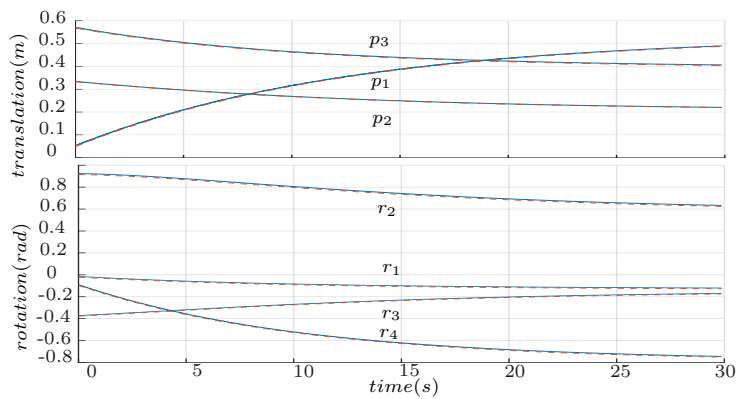%
\par\end{centering}
\caption{
Real-world experiment trajectories with SCLERP-MPC. \textit{Solid blue} curves correspond to measured output, whereas the \textit{Dashed red}  refers to the reference.%
}
\label{fig:rob:trans}
\end{figure}

\begin{figure}
\begin{centering}
\scriptsize
\def\svgwidth{0.7\columnwidth}%{0.94\columnwidth}%
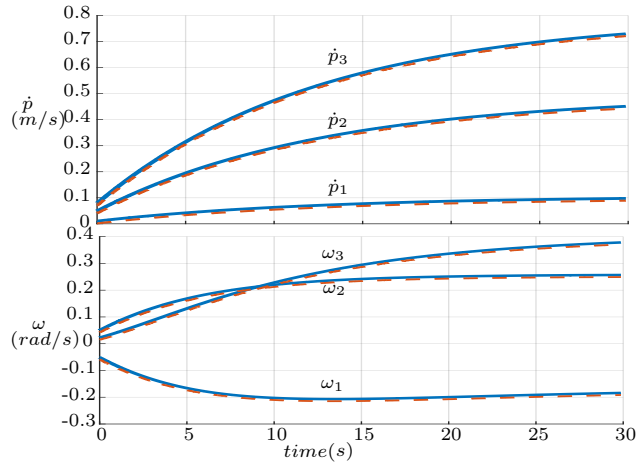%
\par\end{centering}
\caption{Linear and angular components of the end-effector twist for the experiment performed on the real platform. \textit{Solid blue} curves correspond to the values read from the robot, whereas \textit{dashed red} curves to the reference\label{fig:tracking_twist_t}.}
\label{fig:rob:twists}
\end{figure}

Fig.\ref{fig:rob:trans} and Fig.\,\ref{fig:tracking_twist_t} present the translation and rotation components of the pose, and angular and linear components of the end-effector twists successfully tracking the desired twist trajectory.

%%%%%%%figure 1%%%%%%%%
\section{Conclusions and Future Works} \label{conclusion}
This paper presented a cascade structure for the tracking of a smooth coordinate-invariant trajectory using dual quaternion algebra. The proposed architecture integrates a screw-interpolation strategy that satisfies path and geometric constraints within a coordinate-invariant manner with a dual-quaternion algebra MPC that imposes the task-space constraints. The outer-loop MPC performs real-time smoothing of the manipulator's end-effector twist while an inner-loop kinematic controller ensures tracking of the instantaneous desired end-effector pose.

Experiments on a $7$-DoF Franka Emika Panda robotic manipulator have validated the proposed method demonstrating its application to constraint the robot twists, accelerations and jerks within prescribed bounds.

Future works will extend the proposed structure to robot dynamics as well as considerate the inclusion of variable impedance constraints.

%%%%%%%%%%%%%%%%%%%%%%%%%%%%%%%%%%%%%%%%%%%%%%%%%%%%%%%%%%%%%%%%%%%%%%%%%%%%%%%%%%%%%%%%%%%%%%%%%%%%%%%%%%%%%%%%%%%%%%%% APPENDICES %%%%%%%%%%%%%%%%%%%%%%%%%%%%%%%%%%%%%%%%%%%%%%%%%%%%%%%%%%%%%%%%%%%%%%%%%%%%%%%%%%%%%
% \appendix
% 
% \input{mathematical_preliminaries.tex}

%%%%%%%%%%%%%%%%%%%%%%%%%%%%%%%%%%%%%%%%%%%%%%%%%%%%%%%%%%%%%%%%%%%%%%%%%%%%%%%%%%%%%%%%%%%%%%%%%%%%%%%%%%%%%%%%%%%%%%%%%%%%%%%%%%%%%%%%%%%%%%%%%%%%%%%%%%%%%%%%%%%%%%%%%%%%%%%%%%%%%%%%%%%%%%%%%%%%%%%%%%%%%%%%%%%%%%%%%%%%%%%%%%%%%%%%%%%%%%%%%%%%%%%%%%%%%%%%%%%%%%%%%%%%
\addtolength{\textheight}{-2cm}   % This command serves to balance the column lengths

\bibliographystyle{IEEEtran} %plain
\bibliography{biblio}

%%%%%%%%%%%%%%%%%%%%%%%%%%%%%%%%%%%%%%%%%%%%%%%%%%%%%%%%%%%%%%%%%%%%%%%%%%%%%%%%
%\section*{ACKNOWLEDGMENT}

%%%%%%%%%%%%%%%%%%%%%%%%%%%%%%%%%%%%%%%%%%%%%%%%%%%%%%%%%%%%%%%%%%%%%%%%%%%%%%%%

\end{document}